%% file: main.tex
\documentclass{article}
\usepackage{spconf}
\usepackage[utf8]{inputenc}
\newcommand{\hide}[1]{}
\usepackage{xcolor}
\usepackage{url}
\usepackage{hyperref}
\usepackage{mathtools}

\usepackage{hyperref}
\usepackage{url}
\usepackage{graphicx}
\usepackage{graphicx}
\usepackage{subfigure}
\usepackage{amsmath}
\usepackage{amsfonts}
\usepackage{enumitem}
\usepackage{amsthm}
\usepackage[normalem]{ulem}
\useunder{\uline}{\ul}{}

\theoremstyle{definition}

\usepackage{xcolor}
\usepackage{bbm}
\usepackage[utf8]{inputenc}
\usepackage[english]{babel}
\newcommand{\xhdr}[1]{{\noindent\bfseries #1}.}
\newcommand{\cut}[1]{}

\title{Ego-GNNs: Exploiting Ego Structures in Graph Neural Networks}
\name{Dylan Sandfelder, Priyesh Vijayan, and William L. Hamilton\thanks{WLH is a Canada CIFAR Chair in AI. This work was supported by the McGill University SURA Program, as well as Prof. Hamilton's CCAI Chair and a gift grant from Microsoft Research.}}
\address{McGill University, Mila - Quebec AI Institute }

\begin{document}

\maketitle

\input{sections/abstract}
\input{sections/introduction}
\input{sections/related_work}
\input{sections/proposed}
\input{sections/results_table}
\input{sections/theoretical_motive}

\input{sections/results}
\input{sections/conclusion}

\bibliography{refs}
\bibliographystyle{IEEEbib}

\end{document}

%% file: sections/abstract.tex
\begin{abstract}
Graph neural networks (GNNs) have achieved remarkable success as a framework for deep learning on graph-structured data. However, GNNs are fundamentally limited by their tree-structured inductive bias: the WL-subtree kernel formulation bounds the representational capacity of GNNs, and polynomial-time GNNs are provably incapable of recognizing triangles in a graph. In this work, we propose to augment the GNN message-passing operations with information defined on ego graphs (i.e., the induced subgraph surrounding each node). 
We term these approaches Ego-GNNs and show that Ego-GNNs are provably more powerful than standard message-passing GNNs.
In particular, we show that Ego-GNNs are capable of recognizing closed triangles, which is essential given the prominence of transitivity in real-world graphs. We also motivate our approach from the perspective of graph signal processing as a form of multiplex graph convolution. 
Experimental results on node classification using synthetic and real data highlight the achievable performance gains using this approach. 
\end{abstract}


%% file: sections/introduction.tex
\section{Introduction}

Graph neural networks (GNNs) have emerged as a dominant paradigm for representation learning on graph-structured data \cite{hamilton2020grl}.
GNNs have been used to achieve state-of-the-art results for tasks, including predicting the properties of molecules \cite{gilmer2017neural}, forecasting traffic in a large-scale production system\footnote{{https://deepmind.com/blog/article/traffic-prediction-with-advanced-graph-neural-networks}}, classifying protein functions \cite{hamilton2017inductive} and powering social recommendation systems \cite{ying2018graph}. However, despite their empirical successes, GNNs are known to have serious limitations. In particular, the theoretical power of standard GNNs is known to be bounded by the classical Weisfeiler-Lehman (WL) isomorphism test \cite{morris2019weisfeiler, xu2018powerful}.
In signal processing terms, GNNs are known to be limited to simple forms of convolutions \cite{bronstein2017geometric, defferrard2016convolutional}.

One prominent example of the limits of GNNs---and a consequence of their power being bounded by the WL test---is the inability of GNNs to detect the presence of closed triangles in graphs \cite{hamilton2020grl} consistently. This can be proved by showing that, as graph patterns, cycles of length $3$ are not invariant under the color refinement procedure introduced by Arvind et al. \cite{arvind2020weisfeiler}.
In general, GNNs cannot detect the presence of closed triangles, which is a major limitation as transitivity is known to be a critical property in many real-world networks \cite{newman2018networks, watts1998collective}.
In other words, GNNs cannot correctly distinguish the ego-graphs around each node (i.e., the subgraph induced by a node and its immediate neighbors in a graph) since GNNs cannot consistently detect whether two neighbors of a node are connected.

\xhdr{Present work}
We propose an approach to imbue GNNs with information about the ego-structure of graphs explicitly. 
Our approach's basic idea is to combine the standard message-passing used in GNNs with a form of {\em ego-messages}, which are only propagated within the ego-graphs. 
Our approach is theoretically motivated, both in terms of addressing known representational limitations of GNNs, as well as based on multiplex generalizations of graph convolutions. 
Experiments on both real and synthetic data highlight how this approach can alleviate some of the known shortcomings of GNNs. 

%% file: sections/related_work.tex
\section{Related Work}

Our work is closely related to several recent attempts to improve the theoretical capabilities of GNNs.
Early work in this direction focused on elucidating the key properties necessary for a GNN to achieve power equal to basic WL test \cite{morris2019weisfeiler,xu2018powerful}, as well as approaches to design GNNs that can achieve the power of higher-order $k$-WL algorithms \cite{maron2019provably,morris2019weisfeiler}.
Other works have expanded on these ideas, elucidating connections to logic \cite{barcelo2019logical}, dynamic programming \cite{xu2019can}, and statistical learning theory \cite{garg2020generalization}. 

Our work is distinguished from these prior contributions in that we focus on a particular limitation of GNNs, i.e., their inability to exploit ego-structures. 
This limitation is connected to the known representational limits of GNNs: the inability of GNNs to count triangles can be viewed as a consequence of the representational bound from the WL algorithm \cite{morris2019weisfeiler} (or equally as a consequence of GNN's restriction to graded modal logic \cite{barcelo2019logical}).
However, whereas GNNs that are provably more powerful than the WL test in general (e.g., \cite{maron2019provably}) are known to suffer from scalability and stability issues \cite{dwivedi2020benchmarking}, our goal is to provide a targeted theoretical improvement for GNNs, with real-world implications. 

%% file: sections/proposed.tex
\section{Proposed Approach}

Our key proposal is to design a GNN algorithm that can naturally exploit the ego-structure of graphs. In doing so, we hope to improve the theoretical capacity of the model. In addition to that, we hope to potentially address known empirical limitations, such as the {\em over-smoothing} problem that results from node signals becoming uninformative after several rounds of message-passing \cite{hamilton2020grl}.

Our approach involves performing message-passing over the ego-graphs of all the nodes in a graph, rather than simply over the graph itself. 
In the following section, we will formalize how to construct a model that has this desired behavior using a graph, $G$ which is defined by a set of vertices $V$, an adjacency matrix $A \in \mathbb{R}^{|V|\times|V|}$, and a matrix of node features $X \in \mathbb{R}^{|V|\times f}$.

\subsection{Conceptual Motivation}
One way of motivating our Ego-GNN approach is based on the idea of message-passing over a {\em multiplex graph} defined over the original graph. In particular, we construct a multiplex graph $\tilde{G}$ with $|V|$ layers, where it's $i$\textsuperscript{th} layer corresponds to the $i$\textsuperscript{th} node's ego-graph, $\mathcal{G}_i$ from the original graph $G$ and the inter-layer connections are based on the original adjacency structure. Message-passing on such a multiplex would involve passing messages between nodes in the same ego-graph (via the {\em intra-layer} edges), while also propagating information between ego-structures (via the {\em inter-layer} edges).  

If we first define the adjacency matrix of the ego-graph of node $i$, $\mathcal{A}_i \in \mathbb{R}^{|V|\times|V|}$, we can then formally define the multiplex ego-graph $\tilde{G}$ using a {\em supra-adjacency} matrix, $\tilde{A} \in \mathbb{R}^{|V|^2 \times |V|^2}$ constructed from the original graph $G$ as follows:
\begin{equation}\label{eq:supra}
    \tilde{A} = \bigoplus_{i=1}^{|V|}\mathcal{A}_i + (A \otimes I).
\end{equation}
Here, we are creating the {\em intra-layer} adjacency structure by taking the direct sum of 
different adjacency matrices from all the nodes, $i \in V$.
The term $(A \otimes I)$ in Equation \eqref{eq:supra} defines the {\em inter-layer} adjacency structure and is created based on the Kronecker product between the original adjacency matrix and the identity, which creates a group of product graphs \cite{sandryhaila2014big}.
Intuitively, the intra-layer diagonal blocks correspond to each ego-graph, and the off-diagonal blocks connect two ego-graphs if they share a node (i.e., if the corresponding nodes are connected in the original graph). 

Our approach's core idea is to perform message-passing over $\tilde{A}$ rather than the original graph.
However, naively implementing a GNN on the $|V|^2 \times |V|^2$ matrix $\tilde{A}$ would be computationally expensive due to the quadratic increase in the size of the graph. Thus, in the following sections, we describe incremental relaxations and refinements that allow us to implement our Ego-GNN approach in a tractable manner.

\subsection{Sequencing the Intra- and Inter-layer Messages}

In this section, we describe how we approximate message-passing over the full multiplex, defined by the supra-adjacency matrix $\tilde{A}$, using a sequential approach where we run the intra-layer (i.e., within ego-graph) and inter-layer (i.e., between ego-graph) operations in sequence. 
Note also that we assume that our goal is to maintain a single representation $H[i] \in \mathbb{R}^f$ for each node in the graph, and we will use the notation $H_{(t)} \in \mathbb{R}^{|V| \times f}$ to denote the matrix of node representations at layer $t$ of the model. 

\xhdr{Message-passing within ego-graphs}
We first run message-passing independently within each ego-graph. 
To do so, we tile the node representations vertically $|V|$ times before the matrix multiplication:
\[
    \hat{H}_{(t-1)} = \begin{pmatrix} H_{(t-1)} \\ \vdots \\ H_{(t-1)} \end{pmatrix}, \: \hat{H}_{(t-1)} \! \in \mathbb{R}^{|V|^2 \times f}
\]
Multiplying this tiled representation by a power of the intra-layer portion of the supra-adacency matrix, we get 
\begin{equation}
    \hat{H}_{(t)} = \left(\bigoplus_{i=1}^{|V|}\mathcal{A}_i\right)^p\hat{H}_{(t-1)} \label{eq:2},
\end{equation}
where $p$ is the desired scale of the message-passing within each ego-graph (e.g., $p=2$ corresponds to aggregating over two-hop neighborhoods in the ego-graphs). 

\xhdr{Aggregating across ego-graphs}
After Equation (\ref{eq:2}), $\hat{H}_{(t)} \in \mathbb{R}^{|V|^2 \times f}$ contains the representation of every node in every ego-graph.
Therefore, we need to aggregate information across the different ego-graphs to collapse it back into a $|V| \times f$ matrix. 

In our conceptual motivation, we envisioned connected ego-graph layers in the supra-adjacency matrix based on the original adjacency structure of the graph. 
In intuitive terms, we would like the overall representation of each node to depend on its representation in each ego-graph to which it belongs. 
With this in mind, a natural way to approximate the {\em inter-layer} message-passing of the full multiplex is to define each node $i$'s representation in our final matrix $H[i]_{(t)}$ as the average of that node's representation in all of the ego-graphs that contain it. We therefore define each row of $H_{(t)}$ as follows:
\begin{equation}
    H[i]_{(t)} = \frac{ \displaystyle\sum_{j \in Ego(i)}^{} \hat{H}_{(t)}[(j - 1)|V| + i] }{|Ego(i)|} \label{eq:3}
\end{equation}
Where $Ego(i)$ is the list of all the nodes whose ego-graphs contain node $i$.

This approach, based on sequencing the within-ego-graph and between-ego-graph operations, is sound and would produce the desired node representations that leverage the graph's ego-structure. 
However, it has two major flaws. Firstly, it is memory-intensive since $\hat{H}_{(t)}$ cannot be stored using a sparse matrix. Second, the method has not been properly normalized, which can lead to stability issues. 

\subsection{The Ego-GNN Model}

In this section, we build upon the sequential approach proposed above and remedy its key limitations to describe our full Ego-GNN approach. 

\xhdr{Addressing the memory limitations}
The memory problem caused by $\hat{H}_{(t)}$ can be solved by performing the within-ego-graph operations in an iterative manner rather than all at once. 
In particular, instead of carrying out Equation \eqref{eq:2}, we do a combination of operations which calculates $H_{(t)}$ without ever needing to go through $\hat{H}_{(t)}$. These operations are captured in the following sum, which is equivalent to Equation \eqref{eq:3}:
\begin{equation}
    H[i]_{(t)} = \frac{ \left( \displaystyle\sum_{j = 1}^{|V|} \mathcal{A}_j^p H_{(t-1)}  \right) [i]   }{|Ego(i)|} \label{eq:4}
\end{equation}

\xhdr{Adding normalization}
Finally, we only need to apply an appropriate normalization to the model to improve its performance further. 
For generality, we assume that each ego-adjacency matrix $\mathcal{A}_i$ can be replaced by an appropriately normalized counter-part $\hat{\mathcal{A}}_i$ (e.g., the popular symmetric normalization $\hat{\mathcal{A}}_i = D_i^{-\frac{1}{2}}\mathcal{A}_i D_i^{-\frac{1}{2}}$, where $D_i$ is diagonal degree matrix of $\mathcal{A}_i$).

\xhdr{Putting it all together}
Mixing all of these concepts along with the fact that $|Ego(i)|$ is equal to $deg(i) + 1$ when we add self-loops, we can formulate the entire Ego-GNN model with one equation:
\begin{equation}
    H[i]_{(t)} = \frac{ \left( \displaystyle\sum_{j = 1}^{|V|} \hat{\mathcal{A}}_j^p H_{(t-1)}  \right) [i]   }{deg(i) + 1} \label{eq:5}
\end{equation}

\xhdr{Interleaving with standard GNN layers}
Finally, we note that the ego-message passing in Equation (\ref{eq:5}) can be interleaved with layers of standard GNN message-passing, such as the simple propagation rule proposed in \cite{kipf2016semi}.

%% file: sections/results_table.tex
\begin{table*}[!ht]
\centering
\begin{tabular}{llllll}
\multicolumn{6}{c}{\textbf{Node Classification Accuracy}}                                                                                                                                                        \\
\multicolumn{1}{l|}{}                 & \multicolumn{1}{l|}{Cora\cite{mccallum2000automating}}           & \multicolumn{1}{l|}{Citeseer\cite{getoor2005link}}       & \multicolumn{1}{l|}{Pubmed\cite{namata2012query}}         & \multicolumn{1}{l|}{Amazon Computers\cite{shchur2018pitfalls}} & Amazon Photos \cite{shchur2018pitfalls}  \\ \hline
\multicolumn{1}{l|}{\textbf{GCN}}     & \multicolumn{1}{l|}{84.61}          & \multicolumn{1}{l|}{70.88}          & \multicolumn{1}{l|}{\textbf{86.67}} & \multicolumn{1}{l|}{83.32}            & 91.86          \\
\multicolumn{1}{l|}{\textbf{GIN}}     & \multicolumn{1}{l|}{83.76}          & \multicolumn{1}{l|}{69.77}          & \multicolumn{1}{l|}{84.10}          & \multicolumn{1}{l|}{85.27}            & 90.72          \\
\multicolumn{1}{l|}{\textbf{GAT}}     & \multicolumn{1}{l|}{81.86}          & \multicolumn{1}{l|}{69.65}          & \multicolumn{1}{l|}{85.44}          & \multicolumn{1}{l|}{88.61}            & \textbf{92.74} \\ \hline
\multicolumn{1}{l|}{\textbf{Ego-GNN}} & \multicolumn{1}{l|}{\textbf{86.20}} & \multicolumn{1}{l|}{\textbf{72.22}} & \multicolumn{1}{l|}{85.92}          & \multicolumn{1}{l|}{\textbf{89.17}}   & 92.28         
\end{tabular}
\caption{Comparing various methods of performing node classification with the Ego-GNN model.}
\label{tab:accuracy-table}
\end{table*}

%% file: sections/theoretical_motive.tex
\section{Theoretical Motivations}

We briefly remark on some of the theoretical motivations behind the Ego-GNN approach, both in terms of identifying closed triangles as well as motivations from graph signal processing. 

\xhdr{Ego-GNNs and closed triangles}
First, we can note that Ego-GNN layers can be trivially used to recognize the existence of closed triangles in a graph.
For example, assuming constant features as node inputs, counting triangles can be performed via two rounds of message-passing in the ego-graphs: one round to compute node degrees and a second round for the central node to count the degree of each of its neighbors in the ego-graph. 
This simple approach suffices due to the fact that the number of closed triangles in a node's neighborhood is directly computable from the degrees of the nodes in its ego-graph. 
Note, however, that this does not guarantee that an Ego-GNN will learn such a function easily from data. 

\begin{figure}
    \centering
    \includegraphics[width=0.47\textwidth, height=3.9cm]{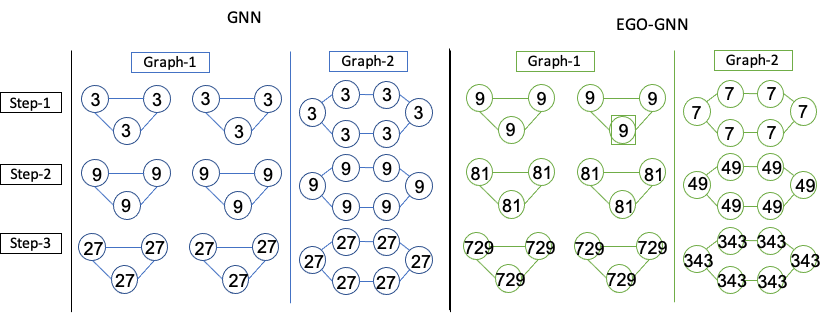}
    \caption{Ego-GNN can distinguish triangles from six-length cycles}
    \label{fig:tri_vs_6C}
\end{figure}

\begin{figure}
    \centering
    \includegraphics[width=0.47\textwidth, height=2.1cm]{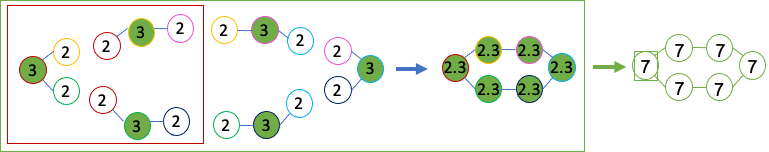}
    \caption{1-step computation of Ego-GNN with p=1}
    \label{fig:ego-compute}
\end{figure}

\xhdr{Ego-GNNs and the classical WL Test}
A critical facet of Ego-GNNs is that they are provably more powerful than the classical 1-WL subtree kernel test. This is especially true when interleaved with standard GNN layers with the same power as the 1-WL subtree kernel. We highlight that our Ego-GNN model can recognize closed triangles with a simple toy problem where the task is to distinguish Graph-1: 3-Cycle ($C_3$) from Graph-2: 6-cycle ($C_6$). Figure \ref{fig:tri_vs_6C} shows outputs of the models on both the graph at different message-passing steps. Unsurprisingly, GNNs are not able to distinguish the two 3-cycles from the 6-cycle graph as the output of every message passing-round is the same. In contrast, the Ego-GNN model distinguished the two graphs by distinguishing wedges and closed triangles in $C_6$ and $C_3$, respectively.
This can be seen in Figure \ref{fig:ego-compute}, which illustrates the first message passing step of Ego-GNN where every node in $C_6$ is able to recognize that their neighbors themselves are not connected within their ego-graph.

\xhdr{Ego-GNNs and graph convolutions}
Ego-GNNs can also be motivated as a graph convolution on the multiplex graph defined by Equation \eqref{eq:supra}. Interestingly, based on the multiplex networks theory, we know that the spectrum of this multiplex graph has a close relationship to the original graph. In particular, based on the perturbation analysis of Sole et al. \cite{sole2013spectral}, we know that there are natural conditions under which the spectrum of this multiplex contains frequencies corresponding to the original graph (via the inter-layer structure) as well as frequencies from the ego-graphs (via the intra-layer structure). 
Moreover, based on the fact that the ego-graphs are all subgraphs of the original graph, the well-known eigenvalue interlacing theorem \cite{chung1997spectral} implies that intra-layer frequencies will interlace the original graph spectrum. This suggests that the Ego-GNN approach can be motivated as a way to give access to a broader set of meaningful frequencies over which to perform graph convolutions.

%% file: sections/results.tex
\section{Experimental results}

In the previous section, we saw that Ego-GNNs have theoretical benefits compared to standard GNNs (e.g., Ego-GNNs can distinguish two triangles from a six-cycle). 
We will now evaluate the empirical performance of Ego-GNNs. 
We first examine a synthetic task in order to demonstrate the ability of Ego-GNNs to reduce the over-smoothing problem, and following this, we examine Ego-GNNs performance on five real-world datasets. 


\subsection{Combating over-smoothing}
Our first experiment shows that Ego-GNNs are capable of effectively combating common over-smoothing problems which crop up regularly with classical GNNs. We can demonstrate this by taking advantage of the stochastic block model (SBM), a type of artificial graph generator which allows its cluster sizes and inter-connectivity probabilities to be pre-specified \cite{holland1983stochastic}. To simulate increased graph signal noise, we gradually increased the inter-cluster connectivity of an artificial SBM graph and measured the node classification performance of the Ego-GNN model, GIN model, and GCN model, where the goal is to classify each node into its underlying community in the SBM graph. 

\begin{figure}[!ht]
  \centering
  \includegraphics[width=\linewidth]{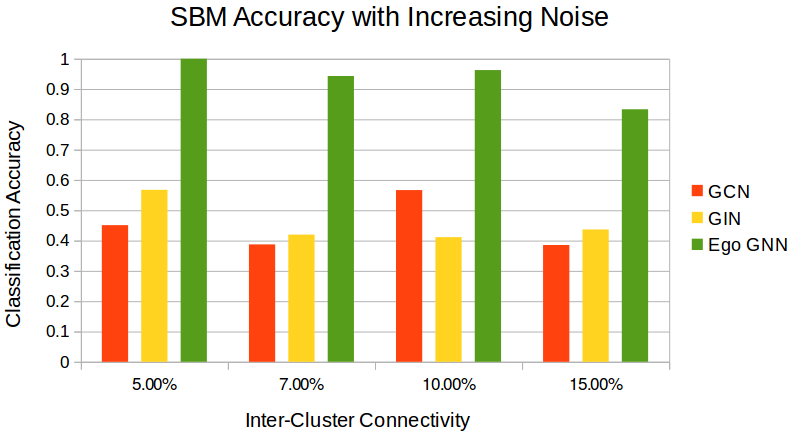}
  \caption{Node classification accuracy results with increasing amounts of edges between the different clusters of a stochastic block model graph.}
  \label{f:1}
\end{figure}

As we can see from the results in Figure \ref{f:1}, the Ego-GNN greatly outperformed the other models even when faced with a mounting density of high-degree nodes connecting disparate clusters. This kind of stability shows how Ego-GNNs can be useful in practice when applied to graphs which are easily susceptible to over-smoothing. \\

\subsection{Classifying nodes on benchmark datasets}
Finally, we compared the performance of the Ego-GNN model with commonly used GNNs in a standard node classification task on five standard and well-studied benchmarks. The results presented in Table \ref{tab:accuracy-table} show that the Ego-GNN model is capable of maintaining a competitive state-of-the-art performance alongside the numerous other useful theoretical properties we have already mentioned in this paper. This makes the Ego-GNN model usable in a wider variety of contexts and applications than other models.

%% file: sections/conclusion.tex
\section{Conclusion}
We have shown that Ego-GNNs are capable of doing more complex graph analysis than other widely used GNN models because of their desirable theoretical properties, specifically their ability to surpass the power of the standard WL test by distinguishing $C_3$ from $C_6$. This result lays the ground-work for further exploration of similar models and presents new ways to test those properties directly.

While constructing deeper convolutional networks in fields like image processing have recently led to large gains in performance, the same can not be said for classical GNNs. Higher order GNNs, like the Ego-GNN model, may carve the path towards overcoming this issue by allowing for the creation of deeper and more scalable graph models that do not suffer from over-smoothing.

%% file: main.bbl
\begin{thebibliography}{10}

\bibitem{hamilton2020grl}
W.~L. Hamilton,
\newblock {\em Graph Representation Learning},
\newblock Morgan and Claypool, 2020.

\bibitem{gilmer2017neural}
J.~Gilmer, S.~Schoenholz, P.~Riley, O.~Vinyals, and G.~Dahl,
\newblock ``Neural message passing for quantum chemistry,''
\newblock in {\em ICML}, 2017.

\bibitem{hamilton2017inductive}
W.L. Hamilton, R.~Ying, and J.~Leskovec,
\newblock ``Inductive representation learning on large graphs,''
\newblock in {\em NeurIPS}, 2017.

\bibitem{ying2018graph}
R.~Ying, R.~He, K.~Chen, P.~Eksombatchai, W.L. Hamilton, and J.~Leskovec,
\newblock ``Graph convolutional neural networks for web-scale recommender
  systems,''
\newblock in {\em KDD}, 2018.

\bibitem{morris2019weisfeiler}
C.~Morris, M.~Ritzert, M.~Fey, W.L. Hamilton, J.~Lenssen, G.~Rattan, and
  M.~Grohe,
\newblock ``Weisfeiler and {L}eman go neural: Higher-order graph neural
  networks,''
\newblock in {\em AAAI}, 2019.

\bibitem{xu2018powerful}
K.~Xu, W.~Hu, J.~Leskovec, and S.~Jegelka,
\newblock ``How powerful are graph neural networks?,''
\newblock in {\em ICLR}, 2019.

\bibitem{bronstein2017geometric}
M.~M. Bronstein, J.~Bruna, Y.~LeCun, A.~Szlam, and P.~Vandergheynst,
\newblock ``Geometric deep learning: Going beyond euclidean data,''
\newblock {\em IEEE Signal Processing Magazine}, vol. 34, no. 4, pp. 18--42,
  2017.

\bibitem{defferrard2016convolutional}
Micha{\"e}l Defferrard, Xavier Bresson, and Pierre Vandergheynst,
\newblock ``Convolutional neural networks on graphs with fast localized
  spectral filtering,''
\newblock in {\em NeruIPS}, 2016, pp. 3844--3852.

\bibitem{arvind2020weisfeiler}
Vikraman Arvind, Frank Fuhlbr{\"u}ck, Johannes K{\"o}bler, and Oleg Verbitsky,
\newblock ``On weisfeiler-leman invariance: subgraph counts and related graph
  properties,''
\newblock {\em Journal of Computer and System Sciences}, 2020.

\bibitem{newman2018networks}
M.~Newman,
\newblock {\em Networks},
\newblock Oxford University Press, 2018.

\bibitem{watts1998collective}
Duncan~J Watts and Steven~H Strogatz,
\newblock ``Collective dynamics of ‘small-world’networks,''
\newblock {\em Nature}, vol. 393, no. 6684, pp. 440--442, 1998.

\bibitem{maron2019provably}
H.~Maron, H.~Ben-Hamu, H.~Serviansky, and Y.~Lipman,
\newblock ``Provably powerful graph networks,''
\newblock in {\em NeurIPS}, 2019.

\bibitem{barcelo2019logical}
Pablo Barcel{\'o}, Egor~V Kostylev, Mikael Monet, Jorge P{\'e}rez, Juan
  Reutter, and Juan~Pablo Silva,
\newblock ``The logical expressiveness of graph neural networks,''
\newblock in {\em ICLR}, 2019.

\bibitem{xu2019can}
Keyulu Xu, Jingling Li, Mozhi Zhang, Simon~S Du, Ken-ichi Kawarabayashi, and
  Stefanie Jegelka,
\newblock ``What can neural networks reason about?,''
\newblock in {\em ICLR}, 2019.

\bibitem{garg2020generalization}
Vikas Garg, Stefanie Jegelka, and Tommi Jaakkola,
\newblock ``Generalization and representational limits of graph neural
  networks,''
\newblock in {\em International Conference on Machine Learning}. PMLR, 2020,
  pp. 3419--3430.

\bibitem{dwivedi2020benchmarking}
Vijay~Prakash Dwivedi, Chaitanya~K Joshi, Thomas Laurent, Yoshua Bengio, and
  Xavier Bresson,
\newblock ``Benchmarking graph neural networks,''
\newblock {\em arXiv preprint arXiv:2003.00982}, 2020.

\bibitem{sandryhaila2014big}
Aliaksei Sandryhaila and Jose~MF Moura,
\newblock ``Big data analysis with signal processing on graphs: Representation
  and processing of massive data sets with irregular structure,''
\newblock {\em IEEE Signal Processing Magazine}, vol. 31, no. 5, pp. 80--90,
  2014.

\bibitem{kipf2016semi}
T.N. Kipf and M.~Welling,
\newblock ``Semi-supervised classification with graph convolutional networks,''
\newblock in {\em ICLR}, 2016.

\bibitem{mccallum2000automating}
Andrew~Kachites McCallum, Kamal Nigam, Jason Rennie, and Kristie Seymore,
\newblock ``Automating the construction of internet portals with machine
  learning,''
\newblock {\em Information Retrieval}, vol. 3, no. 2, pp. 127--163, 2000.

\bibitem{getoor2005link}
Lise Getoor,
\newblock ``Link-based classification,''
\newblock in {\em Advanced methods for knowledge discovery from complex data},
  pp. 189--207. Springer, 2005.

\bibitem{namata2012query}
Galileo Namata, Ben London, Lise Getoor, Bert Huang, and UMD EDU,
\newblock ``Query-driven active surveying for collective classification,''
\newblock in {\em 10th International Workshop on Mining and Learning with
  Graphs}, 2012, vol.~8.

\bibitem{shchur2018pitfalls}
Oleksandr Shchur, Maximilian Mumme, Aleksandar Bojchevski, and Stephan
  G{\"u}nnemann,
\newblock ``Pitfalls of graph neural network evaluation,''
\newblock {\em arXiv preprint arXiv:1811.05868}, 2018.

\bibitem{sole2013spectral}
Albert Sole-Ribalta, Manlio De~Domenico, Nikos~E Kouvaris, Albert Diaz-Guilera,
  Sergio Gomez, and Alex Arenas,
\newblock ``Spectral properties of the laplacian of multiplex networks,''
\newblock {\em Physical Review E}, vol. 88, no. 3, pp. 032807, 2013.

\bibitem{chung1997spectral}
Fan~RK Chung,
\newblock {\em Spectral Graph Theory},
\newblock Number~92. American Mathematical Soc., 1997.

\bibitem{holland1983stochastic}
Paul~W Holland, Kathryn~Blackmond Laskey, and Samuel Leinhardt,
\newblock ``Stochastic blockmodels: First steps,''
\newblock {\em Social networks}, vol. 5, no. 2, pp. 109--137, 1983.

\end{thebibliography}
